\documentclass{article}
\usepackage{spconf,amsmath,graphicx,hyperref}

\usepackage{microtype}
\usepackage{graphicx}
\usepackage{subfigure}
\usepackage{booktabs} 
\usepackage{array} 

\usepackage{tabularx}
\newcolumntype{Y}{>{\centering\arraybackslash}X}

\title{Contextual Range-View Projection for 3D LiDAR Point Clouds
\thanks{This manuscript is a preprint and is currently under consideration for publication in \emph{Computer Vision and Image Understanding}.}
}
\address{School of Innovation, Design and Engineering\\
	Division of Intelligent Future Technologies\\
	Mälardalen University, Västerås, Sweden}
\name{Seyedali Mousavi \qquad Seyedhamidreza Mousavi \qquad Masoud Daneshtalab}
      
\begin{document}
%

\maketitle
%

\begin{abstract}
Range-view projection provides an efficient method for transforming 3D LiDAR point clouds into 2D range image representations, enabling effective processing with 2D deep learning models. However, a major challenge in this projection is the many-to-one conflict, where multiple 3D points are mapped onto the same pixel in the range image, requiring a selection strategy. Existing approaches typically retain the point with the smallest depth (closest to the LiDAR), disregarding semantic relevance and object structure, which leads to the loss of important contextual information.
In this paper, we extend the depth-based selection rule by incorporating contextual information from both instance centers and class labels, introducing two mechanisms: \textit{Centerness-Aware Projection (CAP)} and \textit{Class-Weighted-Aware Projection (CWAP)}. In CAP, point depths are adjusted according to their distance from the instance center, thereby prioritizing central instance points over noisy boundary and background points. In CWAP, object classes are prioritized through user-defined weights, offering flexibility in the projection strategy. Our evaluations on the SemanticKITTI dataset show that CAP preserves more instance points during projection, achieving up to a 4.0 mIoU points improvement over the baseline.
\end{abstract}


\section{Introduction}
\label{sec:intro}
LiDAR sensors have become increasingly prevalent in domains such as autonomous driving, robotics, and mobile mapping~\cite{el2019rgb, krispel2020fuseseg}. These sensors capture spatial information by emitting laser pulses and recording the reflected signals, producing a set of 3D points in Cartesian coordinates that must be processed to support downstream tasks~\cite{qi2017pointnet, milioto2019rangenet++, wu2018squeezeseg, wu2019squeezesegv2}.

Deep learning approaches, including both 3D and 2D models, have emerged as the state of the art for point cloud processing across a wide range of perception tasks. Examples include \textit{semantic segmentation}~\cite{milioto2019rangenet++, ando2023rangevit}, which assigns a semantic label to each point; \textit{panoptic segmentation}~\cite{kirillov2019panoptic}, which combines semantic labeling with instance identification; and \textit{object detection}~\cite{fan2021rangedet}, which involves localizing and classifying objects in a scene.

Point-based networks have been developed to directly process raw point clouds, offering a principled solution for unstructured 3D data~\cite{qi2017pointnet, thomas2019kpconv}. However, their high memory and computational costs limit scalability in large-scale or real-time applications. 
Voxel-based methods discretize space into regular grids to leverage 3D convolutions, improving compatibility with CNNs but suffering from high memory usage and inefficiency caused by data sparsity and the expense of 3D operations~\cite{zhou2018voxelnet}. 
As a result, 3D deep learning methods face significant challenges in meeting the real-time and resource constraints of autonomous driving.~\cite{wu2018squeezeseg, wu2019squeezesegv2}.

To overcome the limitations of point- and voxel-based methods, a widely adopted strategy projects 3D point clouds onto a 2D plane, producing range images that can be efficiently processed with 2D CNNs~\cite{milioto2019rangenet++, ando2023rangevit}. 
This representation not only improves computational efficiency but also provides a regular grid structure that facilitates data fusion, particularly in multi-modal settings such as camera–LiDAR alignment~\cite{el2019rgb, krispel2020fuseseg}.

Nevertheless, range-view projection suffers from inherent issues, most notably the many-to-one point mapping, where multiple 3D points fall into the same pixel of the range image~\cite{kong2023rethinking}. 
While prior work has attempted various refinements, this conflict remains unavoidable due to resolution changes, ego-motion correction, and sensor noise~\cite{triess2020scan, bai2023rangeperception}. 
Existing methods usually keep the point with the smallest depth (closest to the LiDAR), which ensures that visible surfaces are preserved but ignores object structure and semantic importance, leading to the loss of valuable contextual information~\cite{milioto2019rangenet++, ando2023rangevit}.

We address this limitation by enhancing the selection mechanism in range-view projection through the incorporation of contextual information from both instance centers and class labels. Specifically, we introduce Centerness-Aware Projection (CAP), which adjusts point depths according to their distance from the instance center, thereby prioritizing central object points over boundary noise. In addition, we propose Class-Weighted-Aware Projection (CWAP), which leverages user-defined class weights to adapt the projection strategy to specific application requirements.

Extensive experiments on the SemanticKITTI dataset demonstrate the effectiveness of our approach. CAP preserves more instance points during projection and achieves up to a 4.0 mIoU point improvement over baseline methods in semantic segmentation. Moreover, CWAP improves the performance of user-targeted classes through weighted prioritization while maintaining a negligible impact on other classes.

The key contributions of this work are as follows:
\begin{itemize}
    \item \textbf{Centerness-Aware Projection (CAP):} assigns higher priority to points near the centers of object instances, enhancing instance-level range-view projection and improving downstream task performance.
    \item \textbf{Class-Weighted-Aware Projection (CWAP):} incorporates class-dependent weights to emphasize target semantic classes based on user requirements, providing greater flexibility for different applications.
\end{itemize}

The remainder of this paper is organized as follows: Section~\ref{sec:related_work} reviews related work; Section~\ref{sec:review_exist} discusses existing range-view projection methods; Section~\ref{sec:our_methods} introduces the CAP and CWAP approaches; and Section~\ref{sec:experiments} presents experimental results and analysis.

\section{Related Work}
\label{sec:related_work}
LiDAR point cloud processing with deep learning models can be broadly categorized into three groups: \textbf{3D methods}, \textbf{2D methods}, and \textbf{multi-view methods}.
3D methods include point-based approaches, such as PointNet~\cite{qi2017pointnet} and KPConv~\cite{thomas2019kpconv}, which operate directly on raw points, as well as voxel-based approaches, such as VoxelNet~\cite{zhou2018voxelnet}, which discretize space into regular grids for 3D CNNs.
Recent works have also investigated few-shot point cloud semantic segmentation by addressing feature distribution biases between support and query sets, for instance through feature alignment and distribution rectification strategies to improve generalization to unseen classes~\cite{wang2024two}.
While these methods effectively capture fine geometric details, they suffer from substantial memory and computational costs.
2D methods transform point clouds into range image representations via range-view projection, enabling the use of efficient 2D deep learning models. SqueezeSeg~\cite{wu2018squeezeseg} first introduced a \textit{spherical projection}, mapping points into 2D using spherical coordinates and applying CNNs for semantic segmentation. Scan Unfolding~\cite{triess2020scan} later proposed a refined projection strategy that preserves the intrinsic characteristics of LiDAR sensors.
Zou et al.~\cite{zou2019learning} projected LiDAR point clouds onto a cylindrical range map to enable convolutional learning for motion-field estimation, highlighting the applicability of structured range-view representations in learning-based LiDAR processing.
Subsequent works have improved the quality of 2D representations through strategies such as filling missing regions~\cite{chen2024filling} and correcting boundary errors~\cite{bai2023rangeperception}. Recent studies have also explored alternative projection strategies for large-scale point cloud segmentation, including dual projection methods that combine complementary views~\cite{zhao2025dual}. PolarNet~\cite{zhang2020polarnet} introduced a polar grid representation to better align with LiDAR geometry, leading to improved segmentation performance.
Range-view representations have additionally been applied to LiDAR-based object detection~\cite{fan2021rangedet, bai2023rangeperception}. In the context of semantic segmentation, RangeNet++~\cite{milioto2019rangenet++} enhanced model architectures and introduced post-processing techniques, while RangeViT~\cite{ando2023rangevit} leveraged global attention mechanisms to capture long-range dependencies. Kong et al.~\cite{kong2023rethinking} further analyzed design choices that mitigate distortions introduced by range-view projection.
Multi-view methods integrate complementary representations to improve perception accuracy. For example, UniSeg~\cite{liu2023uniseg} combines point-, voxel-, and range-based features, while other approaches aggregate information from multiple views for object detection and segmentation~\cite{chen2017multi, ku2018joint, wang2019pseudo}. In addition, several multi-modal methods fuse LiDAR with RGB data to enhance semantic understanding~\cite{el2019rgb, krispel2020fuseseg, li2023mseg3d, liu2023rgb}. Although effective, these pipelines typically incur higher computational and latency costs.
Given the high memory demands of 3D methods and the increased latency of multi-view and multi-modal approaches, we focus on 2D methods for efficiency and structured representation. In particular, we aim to enhance the range-view projection stage by incorporating contextual information to reduce information loss caused by many-to-one conflicts.

\section{Range-view Projections and Problem Formulation}
\label{sec:review_exist}
\subsection{Spherical Projection and Scan Unfolding}
\begin{figure}[t]
    \centering
    \includegraphics[width=0.8\linewidth]{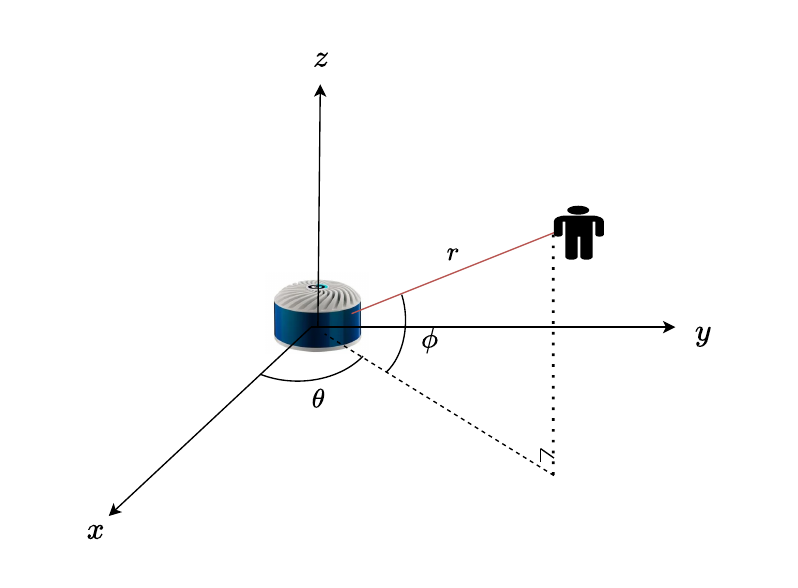}
    \caption{Spherical coordinate representation of a LiDAR measurement, defined by range $r$, azimuth angle $\theta$, and elevation angle $\phi$.}
    \label{fig:lidar_spherical_coords}
\end{figure}

LiDAR sensors operate by emitting laser beams at fixed vertical angles while rotating horizontally. During each 360° rotation, the sensor records a sequence of range measurements corresponding to the reflected laser pulses.  
For each reflected point, the LiDAR records the range $r$ (from time-of-flight) and often the intensity of the return. The corresponding azimuth angle $\theta$ is determined from the horizontal rotation angle, while the elevation angle $\phi$ is fixed by the laser beam’s vertical channel.  
Figure~\ref{fig:lidar_spherical_coords} illustrates the spherical coordinate system used by the LiDAR sensor.
These values are then converted into 3D Cartesian coordinates $(x, y, z)$ using the standard transformation:
\[
x = r \cos{\phi} \cos{\theta}, \quad 
y = r \cos{\phi} \sin{\theta}, \quad 
z = r \sin{\phi}.
\]

Range-view projection methods convert raw LiDAR point clouds into structured 2D range images. In the standard \textbf{spherical projection}~\cite{wu2018squeezeseg}, each point is mapped to image coordinates $(w,h)$ as
\[
\begin{pmatrix}
w \\
h
\end{pmatrix}
=
\begin{pmatrix}
\frac{1}{2} \left( 1 - \frac{\arctan(y/x)}{\pi} \right) W \\
\left( 1 - \frac{\arcsin(z/r) + |f_{\text{down}}|}{f_v} \right) H
\end{pmatrix},
\]
where $W$ and $H$ denote the horizontal and vertical resolutions, respectively, $f_{\text{down}}$ and $f_{\text{up}}$ represent the downward and upward inclination limits of the sensor, and $f_v = |f_{\text{up}}| + |f_{\text{down}}|$ is the vertical field of view. This produces a dense 2D representation aligned with the LiDAR’s field of view.
\textbf{Scan Unfolding}~\cite{triess2020scan} provides a more sensor-faithful alternative. Instead of relying on fixed elevation indices, it reconstructs the range image by mimicking the LiDAR’s acquisition process: at each azimuth step, all beams fire once, producing a single column of the image. By detecting discontinuities in azimuth, Scan Unfolding recovers the scanline ordering and yields a dense projection that closely matches the sensor’s native output, even when such information is not explicitly provided in the dataset.

\subsection{Many-to-One Mapping Problem Formulation}
A key challenge in range-view projection is that multiple 3D points may map to the same pixel, particularly when the horizontal resolution $W$ is small or when the native scanline ordering of the LiDAR is unavailable. This \textbf{many-to-one conflict} leads to ambiguity in which point should represent the pixel. Small $W$ values cause severe information loss (e.g., rasterizing $\sim$120k points into a $64 \times 512$ image retains fewer than 30\% of the points). Increasing $W$ alleviates this issue to some extent but raises computational costs, and conflicts can still arise due to ego-motion correction, noise, or merged scans. 
Several works have attempted to address this problem by adjusting resolution or splitting point clouds into multiple views~\cite{kong2023rethinking}. In contrast, our work focuses on the core question: \emph{when multiple points fall into the same pixel, which one should be selected?}

\section{Proposed Methods}
\label{sec:our_methods}
The standard range-view projection relies on \emph{depth}, selecting the closest point to the LiDAR sensor per pixel. While simple, this approach can fail: dynamic objects (e.g., vehicles, pedestrians) may be ignored in favor of the background, and noisy or unlabeled points (e.g., dust, reflections) may be incorrectly chosen. This indicates that depth-based range-view projection lacks semantic relevance and object structure, resulting in the loss of important contextual information. 

\begin{figure}[t]
    \centering
    \includegraphics[width=0.9\linewidth]{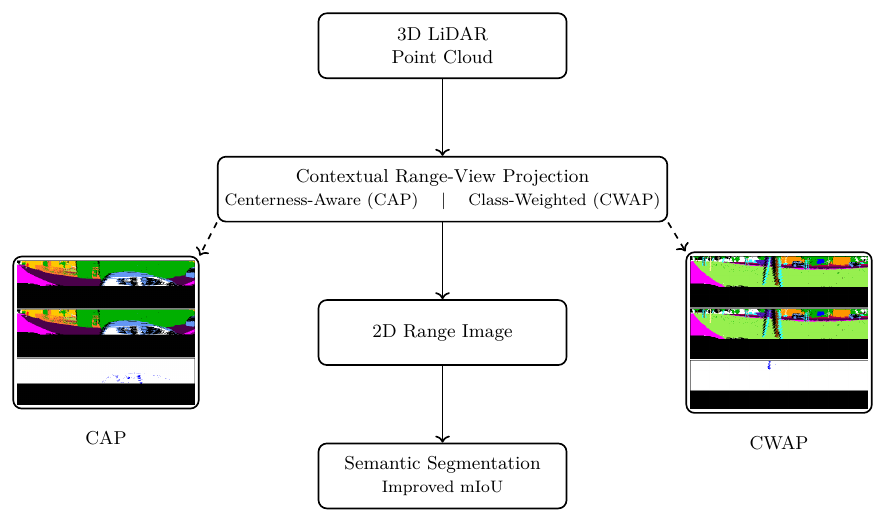}
    \caption{Overview of the proposed contextual range-view projection pipeline. A 3D LiDAR point cloud is projected into a 2D range image using either Centerness-Aware Projection (CAP) or Class-Weighted-Aware Projection (CWAP), which is then processed by a semantic segmentation network to achieve improved performance.}
    \label{fig:method}
\end{figure}

To address this limitation, we propose two contextual range-view projection methods: (1) \textbf{Centerness-Aware Projection (CAP)}, which prioritizes instance-level points, and (2) \textbf{Class-Weighted-Aware Projection (CWAP)}, which incorporates user-specific semantic class weighting. Figure~\ref{fig:method} illustrates our overall pipeline for contextual range-view projection and its integration into a semantic segmentation framework.

\subsection{Centerness-Aware Projection (CAP)}

CAP prioritizes points based on their proximity to the geometric center of object instances during training, such as cars, bicycles, or pedestrians, which are annotated individually and distinguished from background \textit{stuff} categories like roads, buildings, or vegetation.
This encourages the selection of points in range-view projection that better capture the core structure of objects rather than boundary or noisy points.

Given a LiDAR frame represented as an $N \times 4$ matrix, where $N$ is the total number of points, we denote the set of points as $\{\mathbf{p}_i\}_{i=1}^N$ with $\mathbf{p}_i = (x_i, y_i, z_i)$. 
For each object instance, we compute the midpoint of its axis-aligned bounding box as the center point:
{\small
\[
\mu_x = \tfrac{x_{\min}+x_{\max}}{2}, \quad
\mu_y = \tfrac{y_{\min}+y_{\max}}{2}, \quad
\mu_z = \tfrac{z_{\min}+z_{\max}}{2}.
\]
}

Next, CAP quantifies the proximity of each point $\mathbf{p}_i$ to the object center and assigns a centerness score based on a 3D Gaussian distribution:
\[
f(\mathbf{p}_i) = \frac{1}{(2\pi)^{3/2}} 
\exp\!\left(-\tfrac{1}{2}\lVert \mathbf{p}_i-\boldsymbol{\mu}\rVert^2\right),
\]
with $\boldsymbol{\mu} = (\mu_x,\mu_y,\mu_z)$. 
These scores are then normalized to the range $[0,1]$.

The range-view projection criterion in CAP is computed as:
\[
s_i = \|\mathbf{p}_i\|_2 \cdot \frac{1}{f(\mathbf{p}_i)+\varepsilon},
\]
where $\|\mathbf{p}_i\|_2$ is the Euclidean distance of point $i$ to the LiDAR sensor (i.e., depth), and $\varepsilon$ is a small positive constant for numerical stability. For points that do not belong to any instance (stuff points), we set $f(\mathbf{p}_i) = 0$. 

Thus, in CAP, for each pixel in the range image with multiple candidate points, the point with the smallest $s_i$ is selected, preserving the central structure of objects and improving downstream segmentation and localization.

\subsection{Class-Weighted-Aware Projection (CWAP)}
While CAP automatically assigns scores to instances based on their centers, certain applications may require placing greater emphasis on specific instances or background objects. Moreover, since boundary points in CAP receive relatively low scores due to their distance from the center, important object boundaries may be underrepresented or lost.   
To address this, we introduce CWAP, which emphasizes user-specified object classes and can also be applied to non-instance categories (stuff objects).

Given a class $k$, CWAP assigns a weight $\mathbf{w}[k]$ based on the user requirements, which is applied to all its points:
\[
w_i = \mathbf{w}[\text{class}_i] 
\]
Building on this, CWAP range-view projection uses the following weighted criterion for the point selection:
\[
s_i = \|\mathbf{p}_i\|_2 \cdot \frac{1}{w_i+\varepsilon}
\]
Higher weights can reduce $s_i$, thereby increasing the likelihood that the corresponding points are retained. Negative weights can also be assigned (e.g., $w_i=-1$). In this case, if a point with a negative weight competes with points that have positive scores within the same pixel, the negatively weighted point will be selected, and the others will be discarded, regardless of their proximity to the LiDAR. 
This provides flexible control for applications such as emphasizing dynamic agents in forecasting or down-weighting the background in mapping.

\begin{figure}[t]
  \centering
  \includegraphics[width=0.45\textwidth]{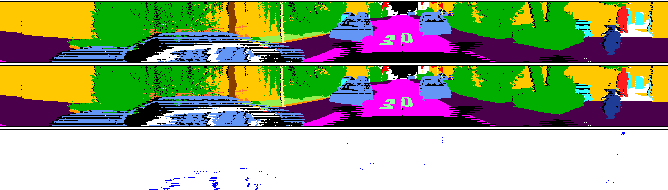}
  \caption{Effect of Centerness-Aware Projection (CAP) on range image formation for a single frame.
  (Top) Standard depth-based projection.
  (Middle) Result of the proposed CAP.
  (Bottom) Pixel-wise differences between the two.
  White pixels indicate unlabeled regions, black pixels correspond to areas with no projected points, and light blue pixels represent cars.}
  \label{fig:range_image_2}
\end{figure}

\begin{figure*}[t]
  \centering
  \includegraphics[width=\textwidth]{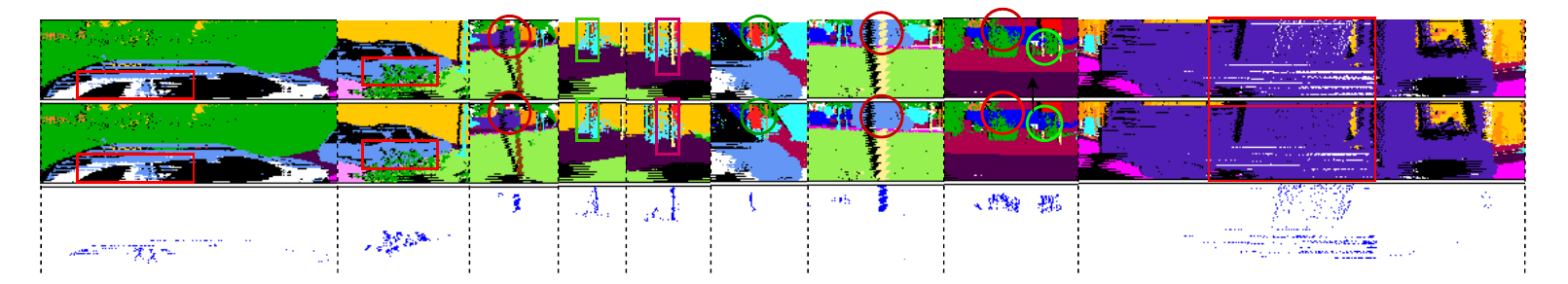}
  \caption{Effect of CAP and CWAP on range image formation for multiple frames.
  (Top) Standard depth-based projection.
  (Middle) Result of the proposed CAP or CWAP.
  (Bottom) Pixel-wise differences between the two.}
  \label{fig:range_image_1}
\end{figure*}

\begin{table*}[t]
\centering
\caption{\textbf{SemanticKITTI} comparison between depth-based, CAP, and CWAP range-view projections methods on the semantic segmentation task.  $\text{mIoU}^{\text{Inst}}$ is the mean over classes 1--8 (Instance classes), and $\text{mIoU}^{\text{Stuff}}$ is the mean over classes 9--19 (Stuff classes). We used the RangeViT architecture for all methods. }
\label{tab:scan_instance_comparison}
\scriptsize
\renewcommand{\arraystretch}{1.}
\setlength{\tabcolsep}{1.0pt}
\resizebox{\textwidth}{!}{
\begin{tabular}{l|*{22}{>{\centering\arraybackslash}p{0.65cm}}}
\toprule
\textbf{Method} &
\rotatebox{90}{car}
& \rotatebox{90}{bicycle}
& \rotatebox{90}{motorcycle}
& \rotatebox{90}{truck}
& \rotatebox{90}{other-veh.}
& \rotatebox{90}{person}
& \rotatebox{90}{bicyclist}
& \rotatebox{90}{motorcyclist}
& \rotatebox{90}{road}
& \rotatebox{90}{parking}
& \rotatebox{90}{sidewalk}
& \rotatebox{90}{other-gr.}
& \rotatebox{90}{building}
& \rotatebox{90}{fence}
& \rotatebox{90}{vegetation}
& \rotatebox{90}{trunk}
& \rotatebox{90}{terrain}
& \rotatebox{90}{pole}
& \rotatebox{90}{t-sign}
& \rotatebox{90}{mIoU}
& \rotatebox{90}{$\text{mIoU}^{\text{Inst}}$}
& \rotatebox{90}{$\text{mIoU}^{\text{Stuff}}$} \\
\midrule

RangeNet++~\cite{milioto2019rangenet++}&
91.4& 25.7& 34.4& 25.7& 23.0& 38.3& 38.8& 4.8& 91.8& 65.0& 75.2& 27.8& 87.4& 58.6& 80.5& 55.1& 64.6& 47.9& 55.9&
52.2 & 35.3 & 64.5 \\
SqueezeSegV3~\cite{xu2020squeezesegv3}&
92.5& 38.7& 36.5& 29.6& 33.0& 45.6& 46.2& 20.1& 91.7& 63.4& 74.8& 26.4& 89.0& 59.4& 82.0& 58.7& 65.4& 49.6& 58.9&
55.9 & 42.8 & 61.7 \\
SalsaNext~\cite{cortinhal2020salsanext}&
91.9& 48.3& 38.6& 38.9& 31.9& 60.2& 59.0& 19.4& 91.7& 63.7& 75.8& 29.1& 90.2& 64.2& 81.8& 63.6& 66.5& 54.3& 62.1&
59.5 & 48.5 & 67.5 \\
KPRNet~\cite{kochanov2007kprnet}&
95.5& 54.1& 47.9& 23.6& 42.6& 65.9& 65.0& 16.5& 93.2& \textbf{73.9}& 80.6& 30.2& 91.7& 68.4& \textbf{85.7}& 69.8& 71.2& 58.7& 64.1&
63.1 & 51.4 & 71.6 \\
RangeViT~\cite{ando2023rangevit}&
95.4 & 55.8 & 43.5 & 29.8 & 42.1 &  63.9 & 58.2 & 38.1 & 93.1 & 70.3 & 80.0 & \textbf{32.5} & \textbf{92.0} & \textbf{69.0} & 85.3 & \textbf{70.6} & \textbf{71.2} & 60.8 & \textbf{64.7} &
64.0 & 53.4 & \textbf{71.8} \\
CAP &
\textbf{95.8} & \textbf{56.2} & 46.7 & 49.1 & \textbf{44.2} & \textbf{64.1} & 64.6 & \textbf{38.2} &
94.6 & 70.2 & \textbf{82.1} & 31.2 & 91.8 & 68.5 & 84.9 & 70.3 & 70.2 & 60.2 & 64.0 &
\textbf{65.8} & \textbf{57.4} & 71.6 \\
CWAP &
93.3 & 35.5 & \textbf{48.5} & \textbf{58.4} & 40.9 & 62.6 & \textbf{55.9} & 38.0 &
\textbf{94.7} & 70.1 & 81.8 & 30.1 & 90.7 & 68.2 & 84.9 & 70.4 & 70.0 & \textbf{61.2} & 64.2 &
64.1 & 54.1 & 71.5 \\
\bottomrule
\end{tabular}
}
\end{table*}

\section{Experiments and Analysis}
\label{sec:experiments} 

\subsection{Datasets and Architecture}
We evaluate our proposed method on range image formation and its impact on overall model performance. We perform experiments on the SemanticKITTI dataset.
To illustrate the effect of our projection strategies, we develop a visualization tool for comparing the resulting range images. To assess the effectiveness of CAP and CWAP in the downstream task, we conduct experiments on semantic segmentation using the RangeViT~\cite{ando2023rangevit} architecture.
The architecture consists of a stem CNN module for feature extraction from range images, a vision transformer for capturing long-range dependencies, and a KPConv classifier for dense prediction. 
To isolate the impact of our projection modifications, we keep the model architecture, training parameters, and all other configurations identical to those in the official RangeViT implementation and only evaluate the effect of range-view projections.

\subsection{Range Image Formation}
To evaluate the quality of our range-view projections, we compare range images generated using the standard depth-based selection with those produced by CAP, and additionally include a third view that highlights pixel-level differences. Fig.~\ref{fig:range_image_2} and Fig.~\ref{fig:range_image_1} illustrate the effect of our methods on range image formation, showing that by leveraging contextual information scoring, CAP (and CWAP) projects more object points into the final range image. For example, in the right part of Fig.~\ref{fig:range_image_1}, the truck instance is represented with noticeably more points when using CAP. The code used for range image formation and comparison is available at \url{https://github.com/s110m/Contextual\_Range-View\_Projection}.

\subsection{Training vs. Inference Usage}
Centerness-Aware Projection (CAP) and Class-Weighted-Aware Projection (CWAP) are utilized only during training, when ground-truth instance and class annotations are accessible. These annotations enable the resolution of many-to-one projection conflicts, improving the quality of the resulting range-view representations and supervision. During inference, such ground-truth information is not available. As a result, the projection reverts to the conventional depth-based approach, assigning each pixel the point closest to the LiDAR sensor. All reported inference results are obtained using this standard projection strategy.

\subsection{Semantic Segmentation}

Table~\ref{tab:scan_instance_comparison} reports the impact of our projection methods (CAP and CWAP) on the semantic segmentation task. CAP demonstrates consistent gains across nearly all instance-level classes, achieving an improvement of 4.0 mIoU points on instance-level categories.
The mIoU for stuff classes decreases by 0.2 mIoU points.
This drop can be attributed to two factors: (1) CAP assigns scores only to instance objects, since center points can be defined for them, and (2) changes in the IoU of certain categories indirectly affect others, as the evaluation metric is sensitive to the balance between true positives and false positives across classes.
For the CWAP strategy, we conduct targeted experiments by assigning a weight of $-1$ to specific object classes (\textit{motorcycle}, \textit{truck}, and \textit{bicyclist}) while setting all others to zero. 
This configuration yields a notable performance gain for the emphasized classes. For instance, performance on the \textit{truck} class improves by 9.3 mIoU points and 28.6 mIoU points compared to CAP and depth-based range-view projection in RangeViT, respectively. In addition, CWAP improves the mIoU on instance-level categories by 0.7 mIoU points compared to depth-based RangeViT method. However, with CWAP, the performance of certain instance objects and stuff classes decreases, as the adjustments made to prioritize specific object categories can create ripple effects on other instances and background classes.

\section{Conclusion}
In this study, we examined depth-based range-view projection for converting 3D point clouds into range image structures. We showed that the point selection strategy plays a critical role due to the many-to-one conflict, affecting both range image quality and downstream perception performance. To address this, we proposed two context-aware training-time projection strategies: Centerness-Aware Projection (CAP), which emphasizes instance-level object structure, and Class-Weighted-Aware Projection (CWAP), which prioritizes task-relevant semantic classes.

Experimental results on SemanticKITTI demonstrate that improving projection quality during training leads to consistent gains at inference, with CAP achieving a 4.0 mIoU point improvement on instance-level categories and CWAP providing targeted gains for selected classes.

\textbf{Future Work.} Future research will focus on estimating centerness or class-aware importance at inference time to enable fully self-contained contextual projection, as well as developing improved class-specific weighting strategies to better handle many-to-one conflicts across a broader range of semantic classes.

\begin{center}
    \textbf{Acknowledgment}
\end{center}

This work was partly supported by the European Union and the Estonian Research Council via project TEM-TA138,
the Swedish Innovation Agency VINNOVA projects AutoDeep and FASTER-AI. The computations were enabled by
resources provided by the National Academic Infrastructure for Supercomputing in Sweden (NAISS), funded by the
Swedish Research Council through grant agreements 2022-06725.

\bibliographystyle{IEEEbib}
\bibliography{strings,refs}

\end{document}